\documentclass[letterpaper, 10 pt, conference]{ieeetrans}  

\IEEEoverridecommandlockouts                              

\usepackage[utf8]{inputenc}
\usepackage{mathtools}
\usepackage{amsmath}
\usepackage{amsfonts}
\usepackage{amssymb}
\usepackage{lipsum}  
\usepackage{xcolor}
\usepackage{graphicx}
\usepackage{booktabs}
\usepackage{hyperref}
\usepackage[sorting = none, backend = bibtex, style=ieee]{biblatex}
\AtNextBibliography{\small}

\newtheorem{remark}{Remark}

\newcommand{\bm}[1]{\boldsymbol{#1}}
\newcommand{\dbm}[1]{\dot{\boldsymbol{#1}}}

\title{\bf Nonlinear MPC for Full-Pose Manipulation of a\\ Cable-Suspended Load using  Multiple UAVs}
\author{Sihao Sun$^{1}$ and Antonio Franchi$^{1,2,3}$ 
\thanks{$^1$ Robotics and Mechatronics Department, Electrical Engineering,  Mathematics, and Computer Science (EEMCS) Faculty, University of Twente, 7500 AE Enschede, The Netherlands. {\tt\footnotesize s.sun@utwente.nl}, {\tt\footnotesize a.franchi@utwente.nl}}
\thanks{$^2$ Department of Computer, Control and Management Engineering, Sapienza University of Rome, 00185 Rome, Italy, {\tt\footnotesize antonio.franchi@uniroma1.it}}
\thanks{$^3$ LAAS-CNRS, Universit\'{e} de Toulouse, 31000 Toulouse, France, {\tt\footnotesize antonio.franchi@laas.fr}}\thanks{This work has been partially funded by the Horizon Europe and H2020 research and
innovation programs under agreement agreement no. 871479 (AERIAL-CORE)}}
\begin{document}

\maketitle

\begin{abstract}
In this work, we propose a centralized control method based on nonlinear model predictive control to let multiple UAVs  manipulate the full pose of an object via cables. At the best of the authors knowledge this is  the first method that takes into account the full nonlinear model of the load-UAV system, and ensures all the feasibility constraints concerning the UAV maximumum and minimum thrusts, the collision avoidance between the UAVs, cables and load, and the tautness and maximum  tension  of the cables. By taking into account the above factors, the proposed control algorithm can fully exploit the performance of UAVs and facilitate the speed of operation. Simulations are conducted to validate the algorithm to achieve fast and safe manipulation of the pose of a rigid-body payload using multiple UAVs.
We demonstrate that the  computational time of the proposed method is sufficiently small ($<$100\,ms)  for UAV teams composed by up to 10 units, which makes it suitable for a huge variety of  future industrial applications, such as autonomous building construction and heavy-load transportation.

\end{abstract}
\section{Introduction}
\label{sec:introduction}

\subsection{Motivation}
A single unmanned aerial vehicle (UAV) has limited load capacities, which impedes its applications in many fields, such as construction and heavy package delivery. 
While increasing the UAV load capacity requires complex mechanical design, using multiple UAVs to transport and manipulate a cable-suspended load is a significantly cheaper and  more promising solution.
However, UAVs are safety-critical systems with limited flying time. 
Existing controllers either neglect safety-related constraints such as the thrust limit of UAVs, or simplifies the model through linearization or quasi-static assumption, limiting their operational speed. 
Therefore, a control framework is needed to simultaneously guarantee safety and agility while manipulating the full pose of a cable-suspended load.

\subsection{Related works}
Manipulating a cable-suspended load using UAVs has been an active research topic in the last decade. 
The simplest case is using a single UAV to control the position of a point-mass load, which has been successfully demonstrated in both simulations and real-world experiments~\cite{ guerrero2015passivity,goodarzi2015geometric,foehn2017fast,son2018model}. 
On the other hand, using multiple UAVs to manipulate an object with cables is still a challenging topic from the control perspective.

\subsubsection{3-DoF manipulation}
In collaborative load transportation problems, multiple UAVs are employed to control the position of a single load. 
Most pieces of research regard the load as a point mass \cite{bernard2011autonomous, lee2013geometric, zurn2016mpc, tartaglione2017model,moreno2018planning,de2019flexible}, with several exceptions using a bar-shape load~\cite{,tagliabue2017collaborative,tognon2018aerial,sundin2022decentralized}. 
To carry a heavy \textit{point mass} load with multiple UAVs, the dynamic coupling effect and kinematic constraints between multiple UAVs and the load needs to be considered. 
One may classify existing methods into centralized controllers which require the states of all UAVs and the load (e.g.,the geometric control~\cite{lee2013geometric}, the model-predictive-control (MPC)~\cite{zurn2016mpc, moreno2018planning}, the distance-based formation-motion control~\cite{de2019flexible}), as well as the decentralized controllers (e.g., the leader-follower controller \cite{gassner2017dynamic, tartaglione2017model}, and distributed MPC controller~\cite{sundin2022decentralized}). 
While the decentralized controller excels in scalability, the centralized methods can utilize more information to achieve better control performance.

\begin{figure}[!t]
    \centering
    \includegraphics[scale=1.0]{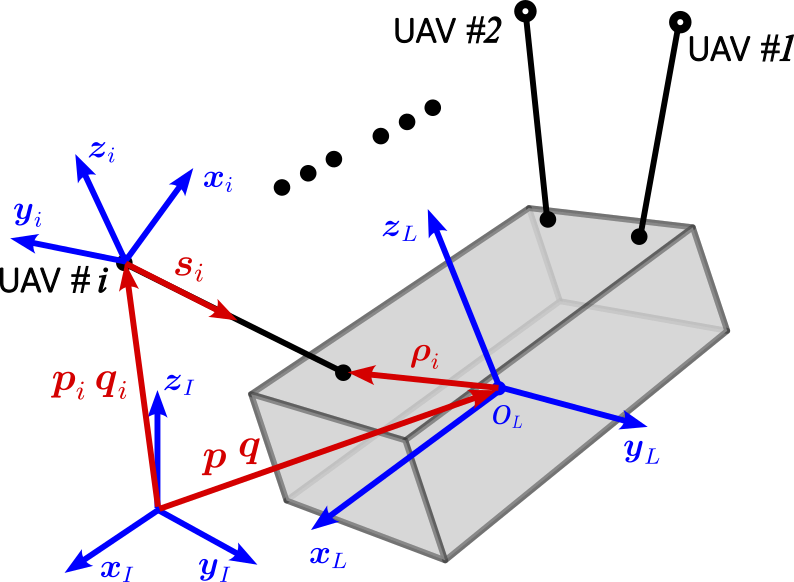}
    \caption{Illustration of multiple UAVs manipulating full-pose of a cable-suspended load, and coordinate systems definition.}
    \label{fig:load_figure}
\end{figure}
\subsubsection{6-DoF manipulation}
When the cable-suspended load is a \textit{rigid body} whose pose (position and attitude) need to be controlled by multiple UAVs, the involvement of the nonlinear rotational dynamic of the load makes the problem more complicated. 
Hence some simplifications in the model are employed in several works.
For instance, the controller in \cite{sanalitro2020full} and sampling-based motion planner proposed in \cite{manubens2013motion} only use kinematic models, which simplifies the design but requires a quasi-static assumption to neglect high-order dynamic effects. 
By contrast, several works also consider the dynamic model of the system, such as the cascaded geometric controller~\cite{lee2017geometric,li2021cooperative}, and the method adapted from cable-driven parallel robots~\cite{sreenath2013dynamics}.
But these reactive controllers cannot address actuator constraints nor avoid collisions between UAVs.

For this reason, MPC can be a promising approach as it can address input and state constraints. 
Its predictive nature can also better exploit the system redundancy and avoid collisions between UAVs. 
A recent work~\cite{wehbeh2020distributed} utilizes linear MPC in both centralized and decentralized fashion to control the full-pose of a rigid body with multiple UAVs. 
However, this method requires linearization of the highly nonlinear model of the system, which has been pointed out to underperform nonlinear MPC (NMPC), especially in highly dynamic tasks~\cite{erunsal2022linear}. 

\subsection{Contribution of this work}
In this work, we propose a centralized NMPC framework to control the full-pose of a cable-suspended load. 
This NMPC uses the nonlinear model of the system and considers safety constraints, including the thrust limitation of each UAV, collision avoidance among UAVs, and guaranteeing non-slack cables. 
Therefore, it can take full advantage of the system's performance to conduct more dynamic manoeuvres while ensuring safety.
We formulate the system into a nonlinear load-cable model, to allow real-time optimization on a laptop.
We have conducted numerical simulations to validate the proposed NMPC to quickly manipulate the pose of a cable-suspended object and comply with all safety-related constraints.

\section{Preliminary}
\label{sec:preliminaries}

\subsection{Problem formulation}
The considered system comprises $n$ UAVs and a single rigid-body load, as is illustrated in Fig.~\ref{fig:load_figure}.
There is a coordinate system $\mathcal{L}=\{\bm{O}_L, \bm{x}_L,\bm{y}_L,\bm{z}_L\}$ attached to the load with the origin located at its centre of mass. 
For each UAV, there is a body-fixed frame $\mathcal{B}_i=\{\bm{O}_{i}, \bm{x}_i, \bm{y}_i, \bm{z}_i\}$ attached where $\bm{O}_{i}$ is at the center of mass of the $i^\textrm{th}$ UAV and $\bm{z}_{i}$ points at the thrust direction. 
The position of the load with respect to the inertial frame $\mathcal{I}$ is denoted by $\bm{p}$, and the rotation from $\mathcal{I}$ to $\mathcal{L}$ is represented by a unit quaternion $\bm{q} = [q_w, q_x, q_y, q_z]^T \in S^3$. 
Similarly, the position of the $i^\textrm{th}$ UAV with respect to $\mathcal{I}$ is denoted by $\bm{p}_i$, and the rotation from $\mathcal{I}$ to $\mathcal{B}_i$ is $\bm{q}_i \in S^3$. 
In addition, we can obtain the rotational matrix with $\boldsymbol{R} = P(\boldsymbol{q})$, where $P:S^3\rightarrow\mathrm{SO(3)}$ is an invertible mapping. 
The quaternion multiplication is represented by matrix multiplication notations, namely $\bm{q}_1\otimes\bm{q}_2=Q(\bm{q_1})\bm{q_2}$, where $\otimes$ indicates the quaternion multiplication, and $Q(\cdot): S^3 \rightarrow \mathbf{R}^{4\times 4}$~\cite{fresk2013full}. 
Throughout the paper, we use bold  lowercase letters to denote vectors, and bold capitalized letters for matrices; otherwise, variables are scalars. 
Vectors expressed in $\mathcal{B}_i$ and $\mathcal{L}$ are written with a superscript, depending on the frame; those expressed in $\mathcal{I}$ have no superscript. 

Each UAV is connected to the load through a cable that connects the centre of mass of the UAV to an arbitrary point on the load. 
Both sides of the cable are connected through universal joints by which only forces are transferred. 
The UAV considered in this work are \textit{Uni-directional Thrust} platforms, whose collective thrust can only vary along one direction (i.e., $\bm{z}_i$). 
Examples are helicopters and coplanar multi-rotor vehicles. 
These platforms can generate moments to change their thrust direction, and consequently, change their position in the inertial frame. 
The objective of the controller is to control the position and attitude (full pose) of the load to track a reference by calculating the desired value of collective thrust and moment of all the UAVs.


\subsection{Modeling}
\subsubsection{Load-cable model}
The load is modelled as a rigid body that complies with the following kinematic and dynamic differential equations:
\begin{align}
    \dbm{p} &= \bm{v}  \label{eq:p_dot}\\
    \dbm{v} &= -\sum_{i=1}^nt_i\bm{s}_i / m + \bm{g} \label{eq:v_dot}\\
    \dbm{q} &= \frac{1}{2}Q(\boldsymbol{q}) 
\left[\begin{array}{c}
     0  \\
     \boldsymbol{\omega}^L 
\end{array}\right]\label{eq:R_dot} \\
    \dbm{\omega}^L &= \bm{I}_L^{-1}\left(-\bm{\omega}^L\times\bm{I}_L\bm{\omega}^L+\sum_{i=1}^nt_i\left(\bm{R}^T\bm{s}_i\times\bm{\rho}_i^L\right)\right) \label{eq:omega_dot}
\end{align}
where $\bm{v}$ and $\bm{\omega}$ respectively represent the linear and angular velocity of the load; $m$ and $\bm{I}_L$ are the mass and the inertia matrix of the load; $\bm{g}$ is the gravity vector. 

For the $i^\textrm{th}$ cable, $t_i$ denotes its tension, $\bm{s}_i\in S^2$ denotes its direction and $\bm{\rho}_i$ is the displacement from $\bm{O}_L$ to the attach point on the load.
We assume the cable is mass-less and non-elastic. Then it is modelled through a high-order kinematic equation up until angular jerks appear, yielding
\begin{align}
    \dbm{s}_i &= \bm{r}_i \times \bm{s}_i \label{eq:cable}\\
    \ddot{\bm{r}}_i &= \bm{c}_i \label{eq:cable_rate}
\end{align}
where $\bm{r}_i$ and $\bm{c_i}$ represent the cable angular velocity and its angular jerk, respectively.
\subsubsection{UAV model}
We model the $i^\textrm{th}$ UAV as a rigid body. The full kinematics and translational dynamics are used in the proposed controller:
\begin{align}
    \dbm{p}_i &= \bm{v}_i \label{eq:pi_dot}\\
    \dbm{v}_i &= (f_i\bm{z}_i + t_i\bm{s}_i)/m_i + \bm{g} \label{eq:vi_dot}\\
    \dbm{q}_i &= \frac{1}{2}Q(\boldsymbol{q}_i)
\left[\begin{array}{c}
     0  \\
     \boldsymbol{\omega}_i^{B_i} 
\end{array}\right] \label{eq:qi_dot}    
\end{align}
where $\bm{v}_i$ and $\bm{\omega}_i$ are linear and angular velocities of the vehicle; $m_i$ and $\bm{I}_i$ are vehicle mass and inertia; $f_i$ is the collective thrust generated by the UAV. 
\subsection{Kinematic constraints}
When cables are taut, the following kinematic constraint holds for the $i^\textrm{th}$ UAV:
\begin{equation}
    \bm{p}_i = \bm{p} + \bm{R}\bm{\rho}_i^L - l_i\bm{s}_i
    \label{eq:p_i}
\end{equation}
where $l_i$ represents the length of the cable attached to the $i^\textrm{th}$ UAV.
Because a (multi-)UAV-load-cable system is differentially flat~\cite{sreenath2013dynamics}, given load position, attitude and cable directions as flat outputs, the states of UAVs can be derived. 
These constraints render the UAV states dependent on the load-cable model.
Note that the controller designed in this paper assumes all cables are taut.
This is guaranteed by including taut cable constraints in the controller design, which will be introduced in Sec.~\ref{subsubsec:tension_constraints}.
\section{Methodology}
\label{sec:methodology}
\begin{figure}
    \centering
    \includegraphics[width=0.48\textwidth]{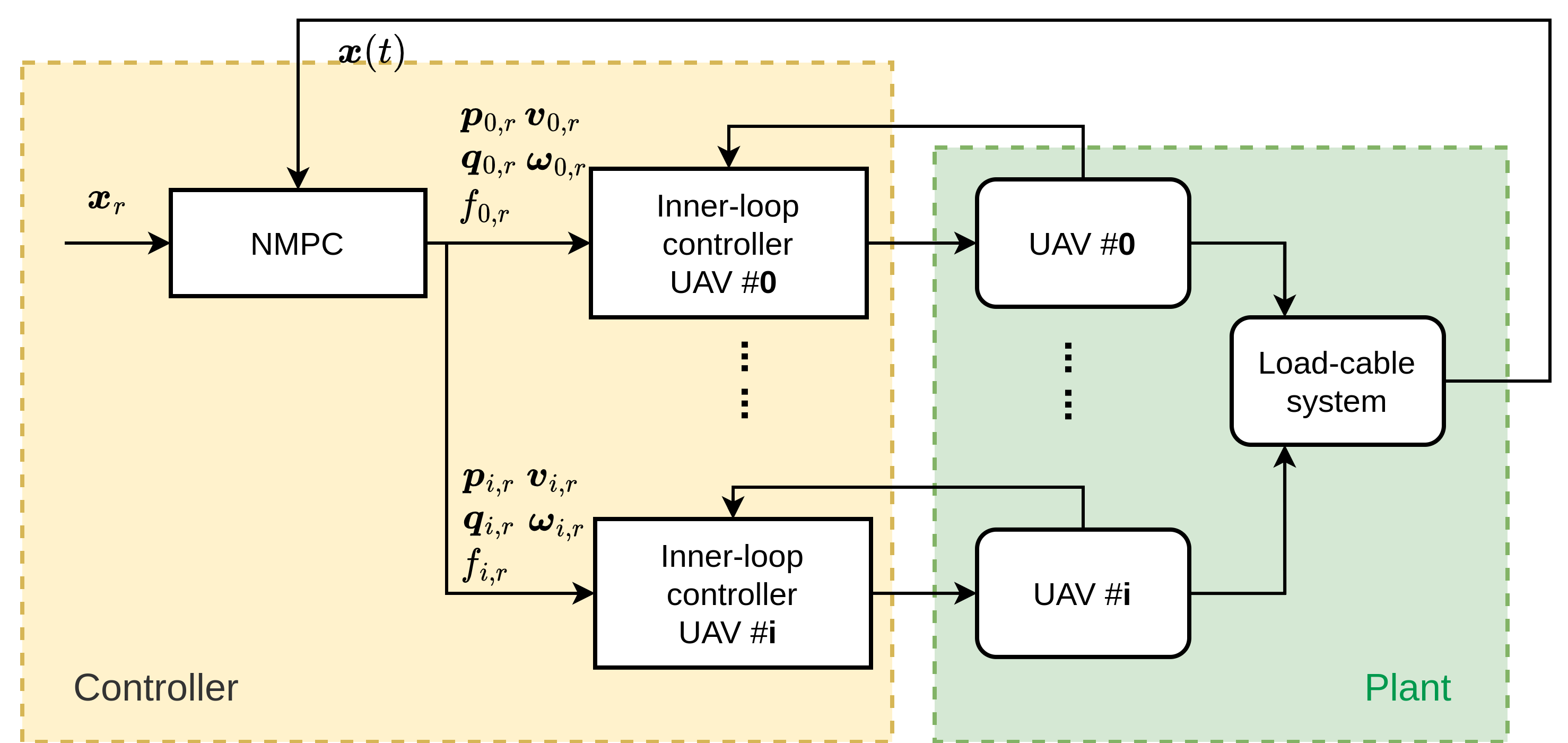}
    \caption{Diagram of the proposed control framework. A centralized NMPC generates references for the inner-loop controller of each UAV.}
    \label{fig:control_diagram}
\end{figure}

This section introduces the proposed control algorithm, including an NMPC as the outer-loop controller and inner-loop controllers for each UAV, as is illustrated in Fig.~\ref{fig:control_diagram}.
The NMPC generates a reference trajectory for each UAV, including $\bm{p}_i$, $\bm{v}_i$, $f_i$, $\bm{q}_i$ and $\bm{\omega}_i$. 
The inner-loop controller is executed at a higher frequency to follow this trajectory until new references are generated by the NMPC.

\subsection{Optimization constraints}
\label{subsec:constraints}
The NMPC solves an optimization problem in a receding fashion. 
For readability, here we first introduce several safety-related constraints to be used in the optimization.
\subsubsection{Thrust constraint}
The NMPC considers the collective thrust limits of each UAV.
By taking the second-order derivative of both sides of (\ref{eq:p_i}) and substituting (\ref{eq:R_dot})-(\ref{eq:vi_dot}) into it, we get the collective thrust of the $i^\textrm{th}$ UAV:
\begin{align}
 f_i\bm{z}_i =\bm{f}_i = &~m_i(\dbm{v} + \bm{R}( \dbm{\omega}^L \times \bm{\rho}_i^L + \bm{\omega}^L\times(\bm{\omega}^L\times\bm{\rho}_i^L))  \notag\\ &- l_i\dbm{r}_i \times \bm{s}_i - l_i\bm{r}_i\times(\bm{r}_i\times\bm{s}_i)))\notag  \\ & - t_i\bm{s}_i-\bm{g} 
 \label{eq:fi_zi}
 \end{align}
 For all UAVs, the thrust constraints are then expressed as
\begin{equation}
    f_{i,min}\leq f_i = \|\bm{f}_i\| \leq f_{i,max},~ \mathrm{for}~i=1,2,...,n
    \label{eq:fi_constraint}
\end{equation}
\subsubsection{UAV minimum distance constraint}
Collisions between UAVs and cables need to be avoided for safety reasons. We use the distance between $i^\textrm{th}$ UAV and the $j^\textrm{th}$ cable as the variable to avoid collisions. This distance is expressed as
\begin{align}
    d_{ij} =& ~\left(\bm{p}_i - \bm{p}_j\right)\times\bm{s}_i \notag\\
    =& ~\bm{R}\left(\bm{\rho}_i^L - \bm{\rho}_j^L\right) + \left(l_j\bm{s}_j-l_i\bm{s}_i\right)\times\bm{s}_i
    \label{eq:dij}
\end{align}
Hence we have distance constraints
\begin{equation}
   0 < d_{min}\leq d_{ij},~\mathrm{for}~i,j=1,2,...,n;~i\neq j
\label{eq:dist_constraint}
\end{equation}
\subsubsection{Non-interference constraint}
A cable should not have any physical interference with the connected UAV and the load. In order to avoid interference between cables and the load, the following constraints are imposed:
\begin{equation}
\bm{s}^L_{i,z} \leq s_{i,z,max} < 0, ~ \mathrm{for}~i=1,2,...,n
\label{eq:slz_constraint}
\end{equation}
where $\bm{s}^L_{i}$ is the $i^\textrm{th}$ cable direction vector expressed in the load frame. These constraints indicate that the cable should be always kept above the connection point on the load.
Similarly, we use the following constraint to ensure that the cable is underneath the connected UAV:
\begin{equation}
    \sigma_i := \bm{f}_i\cdot \bm{s}_i \leq \sigma_{i,max} < 0, ~ \mathrm{for}~i=1,2,...,n
\label{eq:fi_si_constraint}
\end{equation}
\subsubsection{Cable tension constraints}
\label{subsubsec:tension_constraints}
All cables need to be kept taut during the manipulation. And these cables may have maximum permissible tensions. Therefore, we impose the following cable tension constraints:
\begin{equation}
    0 < t_{i,min} \leq t_i \leq t_{i,max}, ~ \mathrm{for}~i=1,2,...,n
    \label{eq:tension_constraint}
\end{equation}

\subsection{Nonlinear model predictive control}
The nonlinear model predictive controller utilizes the load-cable model (\ref{eq:p_dot})-(\ref{eq:cable_rate}). The states and inputs of this model are defined as:
\begin{equation}
\bm{x} = \left[\bm{p},~~\bm{v},~~\bm{q}, ~~\bm{\omega}^L, ~~\bm{s}_1, ~~\bm{r}_1, ~~\dbm{r}_1, ..., ~~\bm{s}_n, ~~\bm{r}_n, ~~\dbm{r}_n \right]    
\end{equation}
\begin{equation}
\bm{u} = \left[\bm{c}_1, ~~ t_1,..., \bm{c}_n, ~~ t_n\right]
\end{equation}
NMPC generates a predicted trajectory in a receding horizon fashion with the horizon length of $h$. 
In each control step of NMPC at time $t$, it solves the following finite-horizon optimal control problem (OCP):
\begin{align}
 \min_{\bm{u}(\cdot),\bm{x}(\cdot)} &~\int_{t}^{t+h} \left(|| \bm{x}_{r}(\tau) - \bm{x}(\tau) ||_{\bm{Q}}^2 + || \bm{u}_{r}(\tau) - \bm{u}(\tau) ||_{\bm{R}}^2\right)d\tau \notag\\
  &~~~~~~~~+|| \bm{x}_{r}(t+h) - \bm{x}(t+h) ||_{\bm{Q}_e}^2 \notag \\
\text{s.t. }~& \textit{dynamic constraints:}\notag\\
        &~~~~~~~~(\ref{eq:p_dot})-(\ref{eq:omega_dot}) \notag\\
        & ~~~~~~~~(\ref{eq:cable})(\ref{eq:cable_rate})~~\textbf{for}~i~=~1,\ldots,n \notag\\
        & \textit{inequality path constraints:}\notag\\
        &~~~~~~~~(\ref{eq:fi_constraint})(\ref{eq:dist_constraint})(\ref{eq:slz_constraint})(\ref{eq:fi_si_constraint})(\ref{eq:tension_constraint})\notag
\end{align}
The above nonlinear OCP is discretized with a multi-shooting approach~\cite{bock1984multiple}. 
The states and inputs over the time horizon $\tau \in [t, t+h]$ are discretized into $N$ equal intervals with step size $\Delta t = h/N$. 
The inequality path constraints are formulated as soft constraints with slack variables. 
Then the OCP is transformed into a nonlinear programming problem, which is solved by sequential quadratic programming (SQP) in a real-time iteration (RTI) scheme~\cite{diehl2002real}. The SQP is solved with HPIPM~\cite{frison2020hpipm} as the quadratic-programming solver.
All the above procedures are implemented using the ACADOS toolkit~\cite{verschueren2022acados}.

By solving the above OCP, the optimal states and control along the discrete nodes are obtained. 
Given the state and input at the $k^\textrm{th}$ node on the predicted trajectory, denoted by $\bm{x}|_k$ and $\bm{u}|_k$, we can get the position, velocity, and acceleration of each UAV using the kinematic constraint (\ref{eq:p_i}) and its derivatives. 
One step further, we can also get the thrust and thrust direction through (\ref{eq:fi_zi}). They are denoted by $f_i|_k$ and $\bm{z}_i|_k$ at the $k^\textrm{th}$ node. 

Note that the heading of each UAV is independent of the thrust direction dominating the forces acting on the load. Therefore, the reference heading can be selected independently for other purposes such as achieving better perception accuracy \cite{falanga2018pampc}. After defining the reference heading of the $i^\textrm{th}$ UAV as $\psi_{i,r}$, we can calculate the reference attitude of the $i^\textrm{th}$ UAV at the $k^\textrm{th}$ node in the quaternion form through following steps:
\begin{align}
\boldsymbol{e}_{i}|_k &= [\cos{\psi_{i,r}},~\sin{\psi_{i,r}},~0]^T, \label{eq:e_i_k}\\
\boldsymbol{y}_{i}|_k &= \frac{\boldsymbol{z}_{i}|_k\times\boldsymbol{e}_{i}|_k}{||\boldsymbol{z}_{i}|_k\times\boldsymbol{e}_{i}|_k||}, \\
\boldsymbol{x}_{i}|_k &=\boldsymbol{y}_{i}|_k \times \boldsymbol{z}_{i}|_k, \\
\boldsymbol{q}_i|_k &= P^{-1}\left(\left[ \boldsymbol{x}_i|_k,~\boldsymbol{y}_i|_k,~\boldsymbol{z}_i|_k\right]\right) \label{eq:q_i_reference}
\end{align}
where $P^{-1}(\cdot)$ converts the rotation matrix to quaternion.

Then the angular rate of the $i^\textrm{th}$ UAV along the trajectory can be calculated through (\ref{eq:qi_dot}). Specifically, we use finite difference with the trapezoidal rule, yielding
\begin{equation}
\left[  \begin{array}{c}
         0 \\
         \bm{\omega}_i|_k
    \end{array}\right]
     = \frac{4}{\Delta t} \left(Q(\bm{q}_i|_{k+1}) + Q(\bm{q}_i|_{k})\right)^{-1} \left(\bm{q}_i|_{k+1} - \bm{q}_i|_{k} \right)
    \label{eq:omega_reference}     
\end{equation}

\begin{remark}
Cable angular jerk $\bm{c}_i$ has the same relative degree as the UAV body rate $\bm{\omega}_i$ with respect to load pose. Therefore, we set cable angular jerk $\bm{c}_i$ as the input of the NMPC and apply constraints on it, which can ensure a bounded UAV body rate reference and subsequently improve the smoothness of the predicted UAV trajectory. 
\end{remark}

\begin{remark}
The cable tension $t_i$ is regarded as input of the OCP rather than state variables.
Thus they are not needed to be measured.
\end{remark}

\subsection{Inner-loop controller of each UAV}
Now we have reference states of all UAVs along the predicted trajectory, which can be tracked by various approaches, such as flatness-based controllers (e.g.,~\cite{tal2020accurate}) and model predictive control~\cite{sun2022comparative}. We refer the readers to these pieces of literature for details. Our simulation tests indicate that a 10~Hz of update rate from NMPC is sufficient for this scheme. Another possible scheme is to directly feed the collective thrust, attitude and angular rate along the predicted trajectory to an inner-loop attitude controller (e.g., the geometric controller~\cite{lee2010geometric}), as long as the NMPC update rate is sufficiently high (empirically, approx. $>$100Hz). The selection between these two schemes depends on the update rate of NMPC determined by the computational budget and the number of UAVs. Uncertainties in the dynamic model such as complex aerodynamic effects~\cite{sun2019aerodynamic} also favor a higher update rate from the NMPC. We will particularly investigate the effect of model uncertainties on our approach in future work.

\section{Numerical Validation}
\subsection{Simulation setup}

The numerical validation is conducted in a simulator provided by ACADOS toolkit using the load-cable model (\ref{eq:p_dot})-(\ref{eq:cable_rate}), and indirectly obtain the UAV states using (\ref{eq:p_i}) and (\ref{eq:e_i_k})-(\ref{eq:omega_reference}).
The weights of the NMPC are given in Table~\ref{tab:weights}. Note that terminal weights $\bm{Q}_e$ are selected identical to $\bm{Q}$.

\begin{figure}[t]
    \centering
    \includegraphics[width=0.48\textwidth, trim={6cm 5cm 4.0cm 5cm},clip]{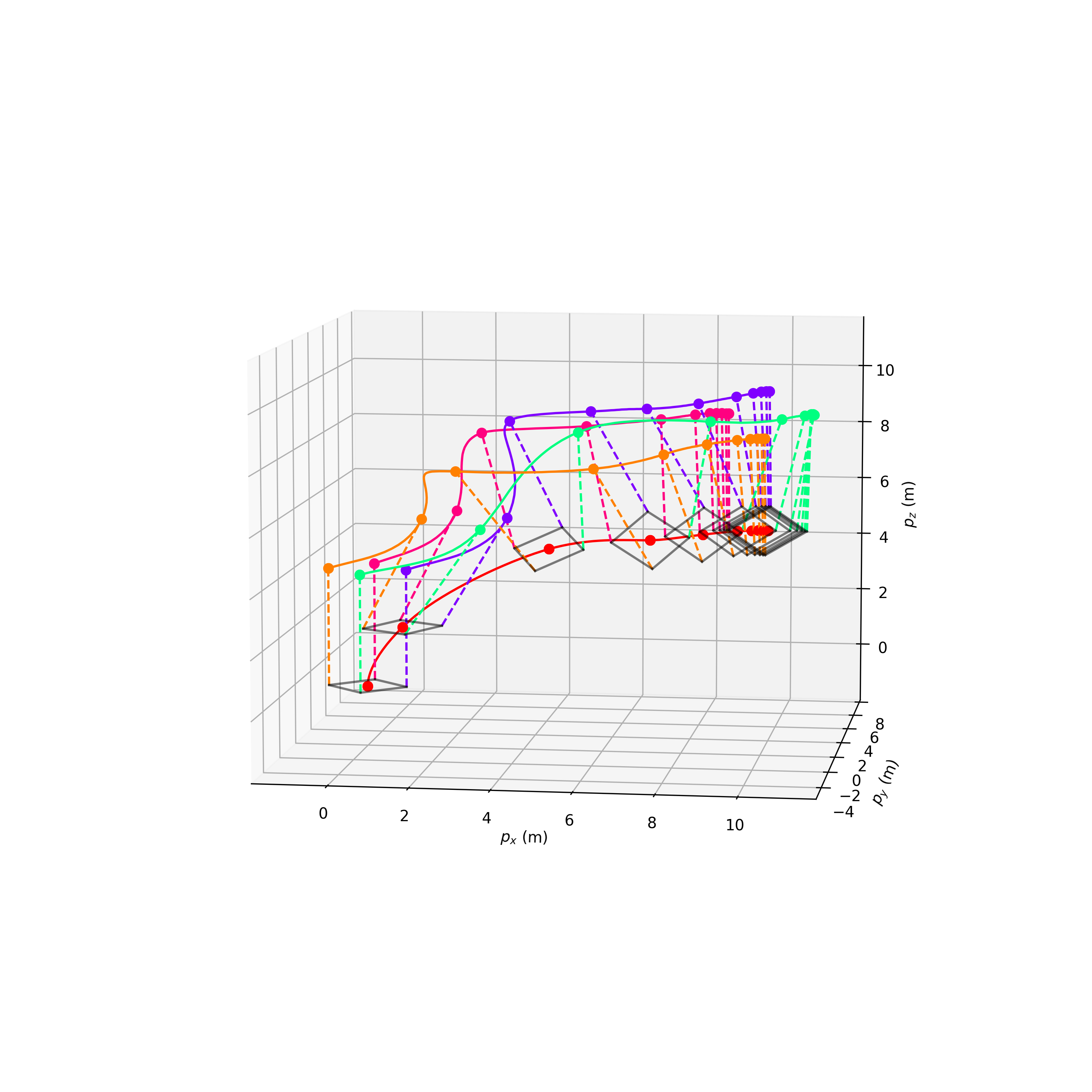}
    \caption{3D path of the load and UAVs in the inertial frame. The pose of the load are plotted \textit{every second} in dark grey. Solid-red line represents the centre of mass of the load. The position of UAV \#1 to \#4 are plotted respectively in magenta, purple, green, and orange. Dashed lines represent the cables.}
    \label{fig:3d_plot}
\end{figure}

As an example, we demonstrate a case using four UAVs to manipulate a rigid-body load. 
The load mass is 4\,kg, and each UAV weighs 1\,kg. 
The length of cables connecting each UAV and the load are all 4\,m. 
The four connection points on the load are evenly distributed every 90 degrees on the horizontal plane, all ranging 1\,m from $\bm{O}_L$.
The initial position of the load is $\bm{p}=[0,~0,~0]$~m, and the initial attitude is $\bm{q}=[0,~0,~0,~1]$. 
Then we send a constant reference position of $\bm{p}_{r}=[10,~3,~5]$\,m, and a constant reference attitude of $\bm{q}_{r}=[0.27, -0.27, 0.65, 0.65]$ (i.e., 90 degrees yaw angle and 45 degrees roll angle) are sent to the controller.
For all four cables, their tension references are set as 9.8~N. The remaining variables in $\bm{x}_{r}$ and $\bm{u}_{r}$ are set as zero.
Note that in this simulation, a constant terminal set point is used as the reference, despite that the method also allows following a time-varying trajectory.

\begin{table}[b]\centering
\caption{Weights of the NMPC.}\label{tab:weights}
\scriptsize
\begin{tabular}{l|l}\toprule
Variable & Values \\\midrule
$\bm{Q}_p$ & diag(200, 200, 200) \\
$\bm{Q}_v$ & diag(200, 200, 200) \\
$\bm{Q}_q$ & diag(500, 500, 500) \\
$\bm{Q}_w$ & diag(50, 50, 50)\\
$\bm{Q}_{s,i}$ & diag(50, 50, 50) \\
$\bm{Q}_{r,i}$ & diag(100, 100, 100) \\
$\bm{Q}_{\dot{r},i}$ & diag(10, 10, 10) \\
$\bm{R}_{c,i}$ & diag(1, 1, 1) \\
$\bm{R}_{t,i}$ & diag(1, 1, 1) \\
\bottomrule
\end{tabular}
\end{table}

\subsection{Simulation results}
\label{sec:results}

\begin{figure}[t]
    \centering
    \includegraphics[trim={0cm 0cm 0.0cm 0.5cm},width=0.48\textwidth]{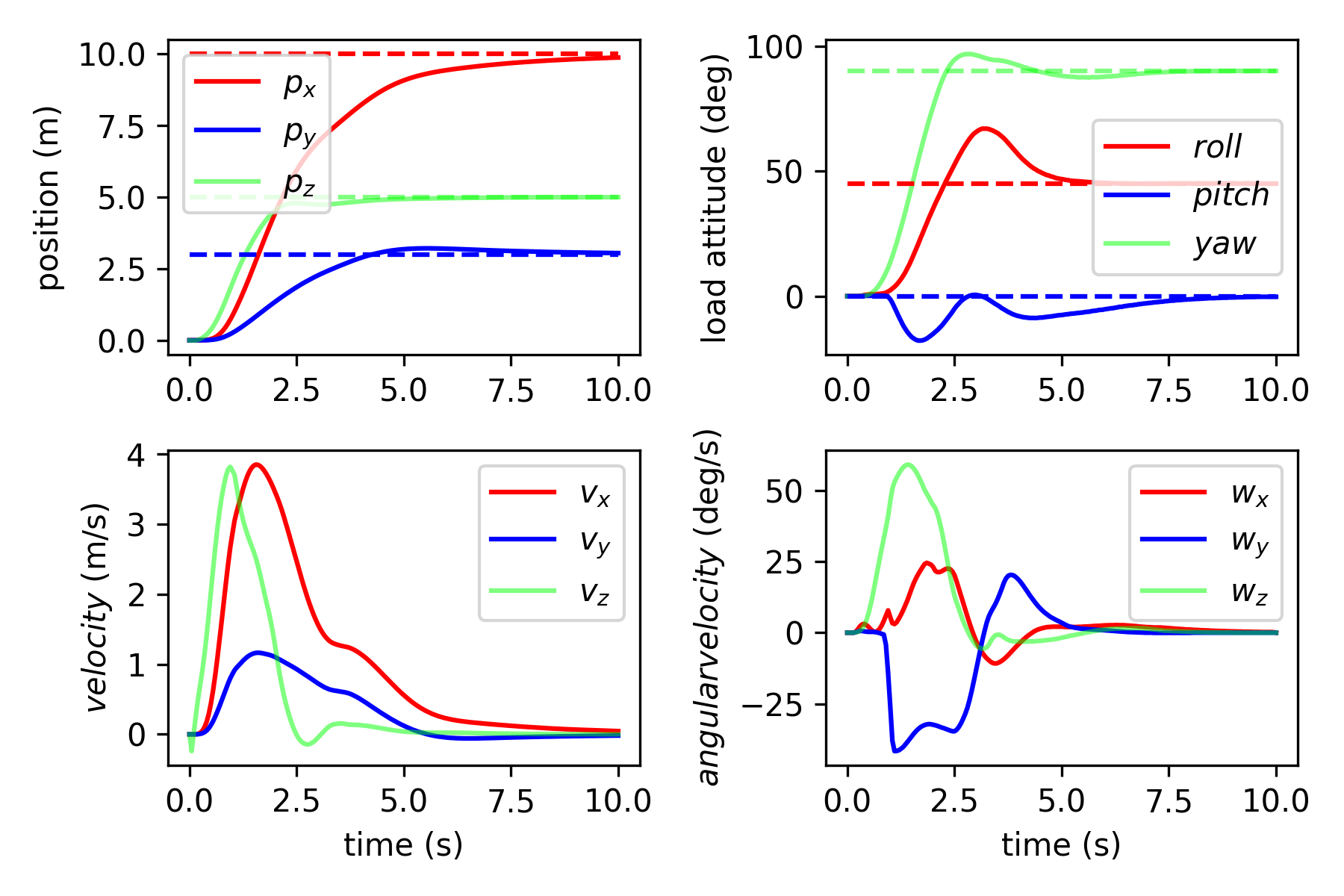}
    \caption{Time history of the position, orientation, velocity, and angular rate of the load being manipulated by 4 UAVs. Dash lines in these figures represent the reference.}
    \label{fig:trajectory}
\end{figure}
\begin{figure}[h!]
    \centering
    \includegraphics[trim={0cm 0cm 1.0cm 0.7cm}, width=0.48\textwidth]{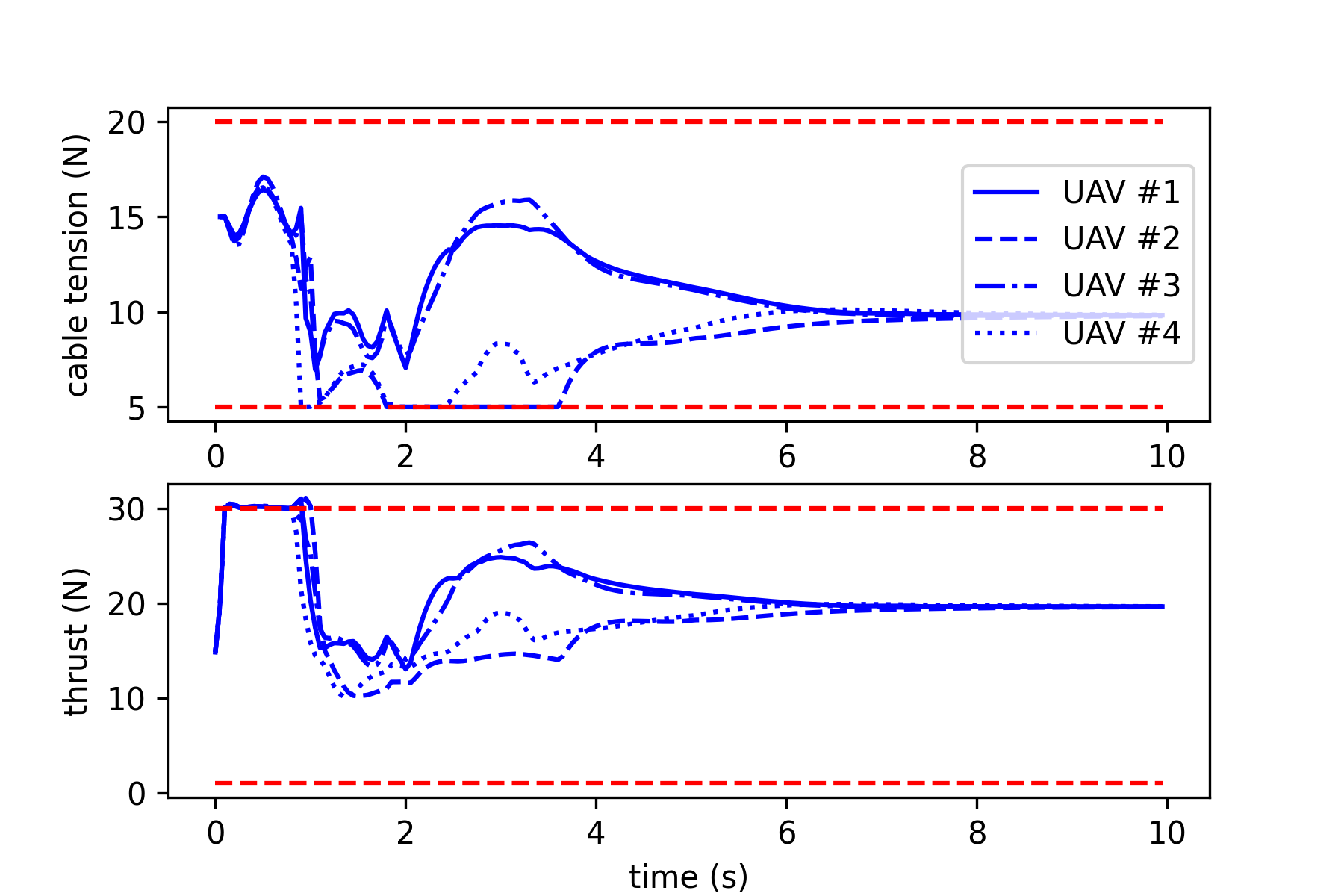}
    \caption{Time history of the cable tension and the collective thrust of all UAVs. Red-dash lines represent the constraints.}
    \label{fig:constraints_t_thrust}
\end{figure}

\begin{figure}[t!]
    \centering
    \includegraphics[trim={0cm 0cm 1.0cm 1.3cm},width=0.48\textwidth]{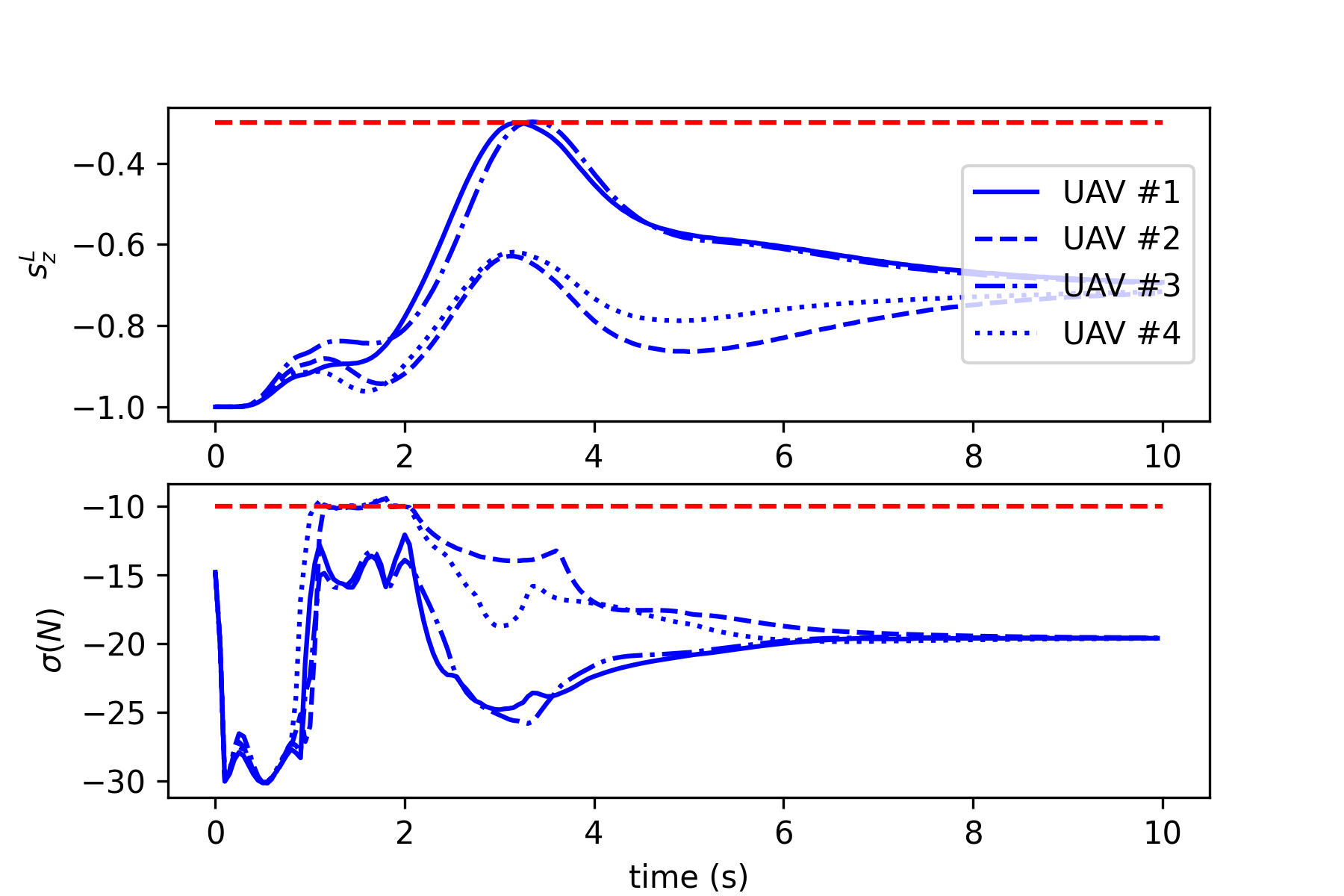}
    \caption{Time history of variables related to interference between cables and UAVs (i.e., $\sigma_i$), and between cables and the load ($\bm{s}^L_z$). Red-dash lines represent the constraints ($s_{i,z,max}=-0.3,~\sigma_{i,max}=-10$\,N)}.
    \label{fig:constraints_interference}
\end{figure}

\begin{figure}[t]
    \centering
    \includegraphics[trim={0cm 0cm 1.0cm 0cm},width=0.48\textwidth]{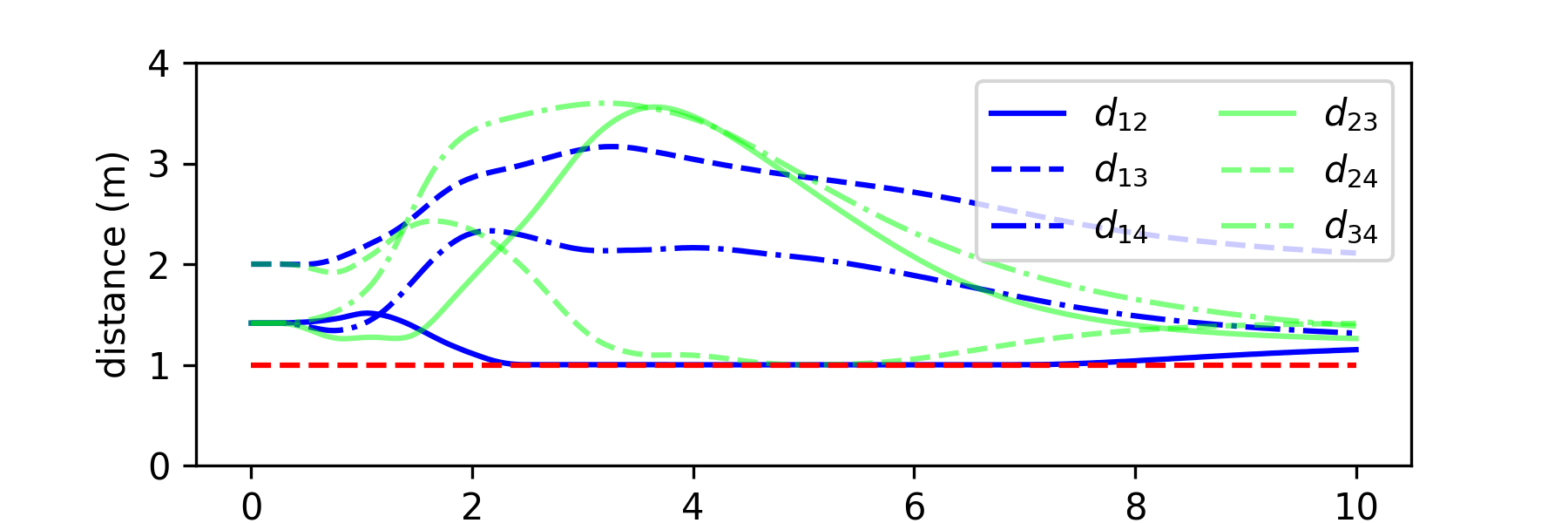}
    \caption{Time history of the relative distances between UAVs. Red-dash lines represent the minimum distance (1.0\,m) set in the controller.}
    \label{fig:constraints_dist}
\end{figure}

Fig.~\ref{fig:3d_plot} presents the 3D path of the load and the UAVs, as well as the attitude of the load along the trajectory. 
Fig.~\ref{fig:trajectory} presents the time history of the pose, velocity, and angular rate of the load. 
The pose errors converge to zero within 10 seconds.  
The maximum horizontal and vertical velocity are over 4\,m/s for a traversal of 10\,m, which is far more agile than the existing approach(e.g, ~\cite{wehbeh2020distributed,lee2017geometric}).
Hence the pose errors have been reduced by over 80\% in the first 5 seconds.

Fig.~\ref{fig:constraints_t_thrust} plots the time history of the cable tension, as well as the thrust generated by each UAV. 
In the initial phase of the tracking procedure, four UAVs jointly increase the altitude of the load. 
Meanwhile, the thrusts of each UAV comply with their maximum thrust constraints.
Shortly after the load reaches the maximum vertical velocity at around 1\,s, the UAVs start to reduce the tension of the cable to decelerate the load.
The proposed NMPC keeps the cable tension above the minimum value, which is set as 5\,N to prevent cable slack.

Variables related to non-interference constraints are plotted in Fig.~\ref{fig:constraints_interference}, which includes interference between cables and the load indicated by $\bm{s}^L_{i,z}$, and the interference between cables and UAVs indicated by $\sigma_i$ defined in (\ref{eq:fi_si_constraint}). 
It is possible to see how all the variables comply with the constraints. 
Fig.~\ref{fig:constraints_dist} presents the distance between UAVs and cables of other UAVs. 
During the task, all these distances are larger than a threshold, set as 1\,m, to avoid collision between UAVs and cables. 

\begin{figure}[t]
    \centering
    \includegraphics[trim={0cm 0cm 1.0cm 0.8cm},width=0.48\textwidth]{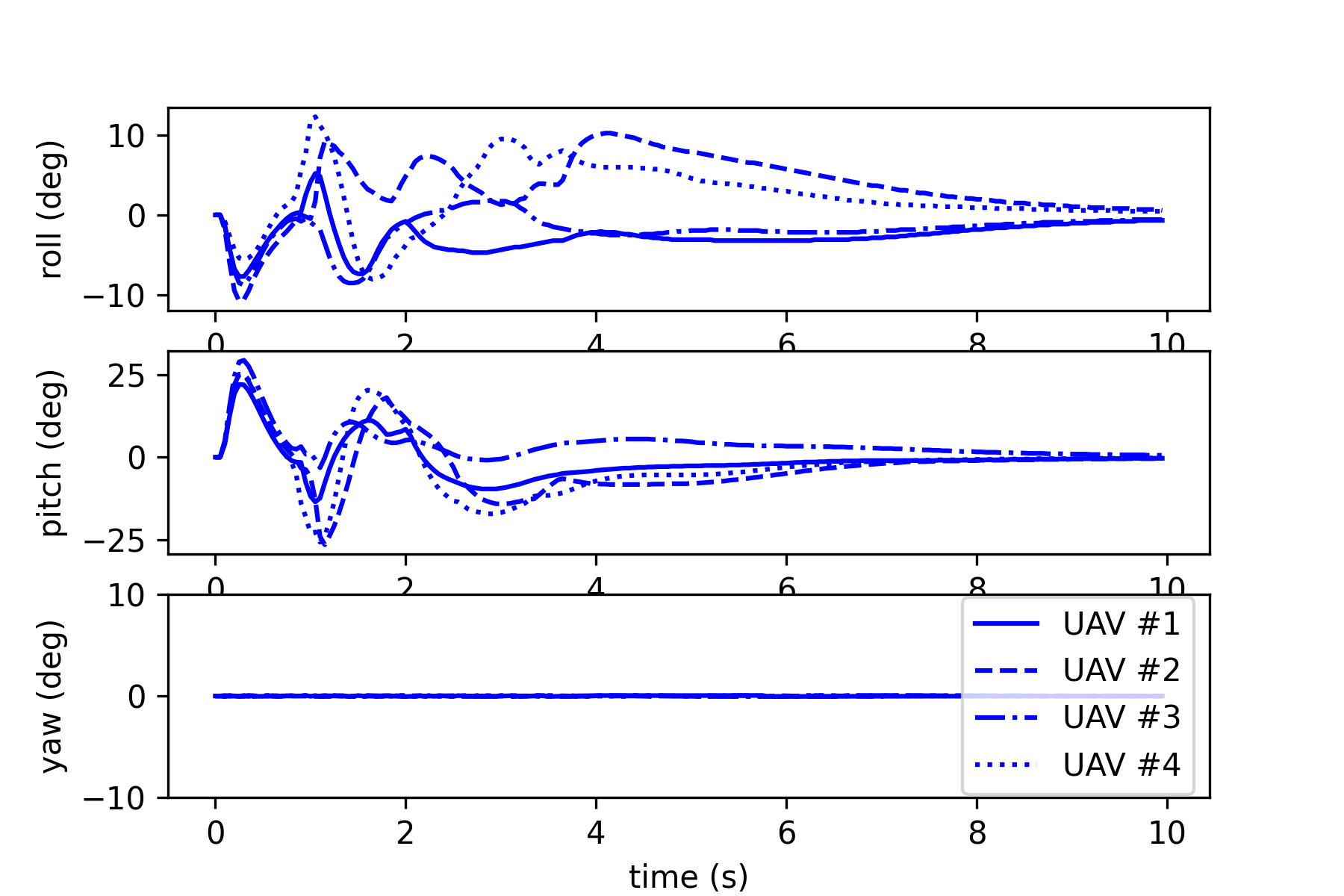}
    \caption{Time history of the attitude of UAVs.}
    \label{fig:attitude_UAV}
\end{figure}

\begin{figure}[t]
    \centering
    \includegraphics[trim={0cm 0cm 1.0cm 0cm},width=0.48\textwidth]{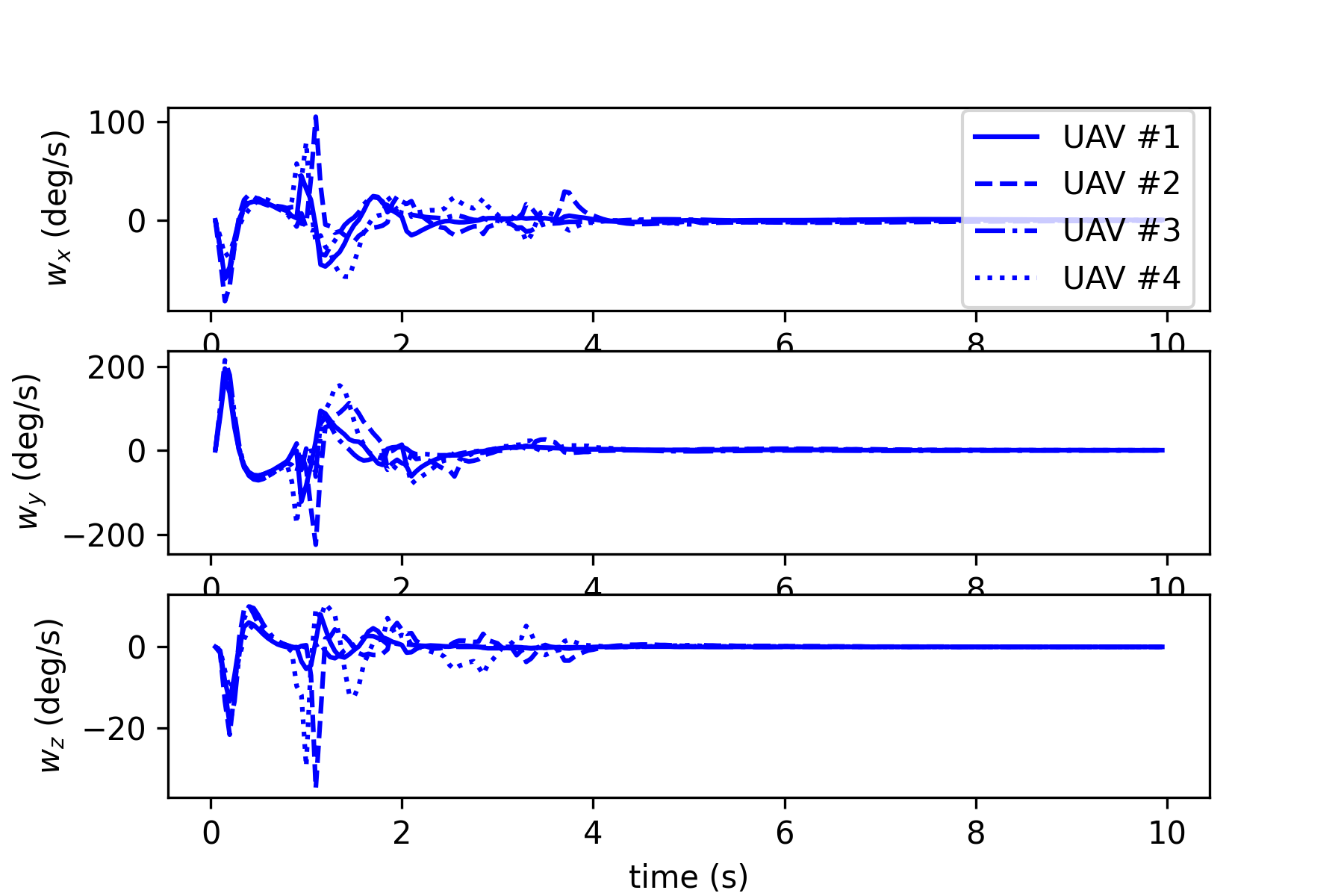}
    \caption{Time history of the angular velocities of UAVs}
    \label{fig:body_rate_UAV}
\end{figure}

Fig.~\ref{fig:attitude_UAV} and Fig.~\ref{fig:body_rate_UAV} respectively show the attitude and the angular velocity of all the UAVs.
The attitude is plotted in Euler angles for better intuition ('zyx' convention).
Note that the reference heading angle of all UAVs are set as zero in this simulation.

\subsection{Computational time}
\begin{figure}[t]
    \centering
    \includegraphics[width=0.4\textwidth]{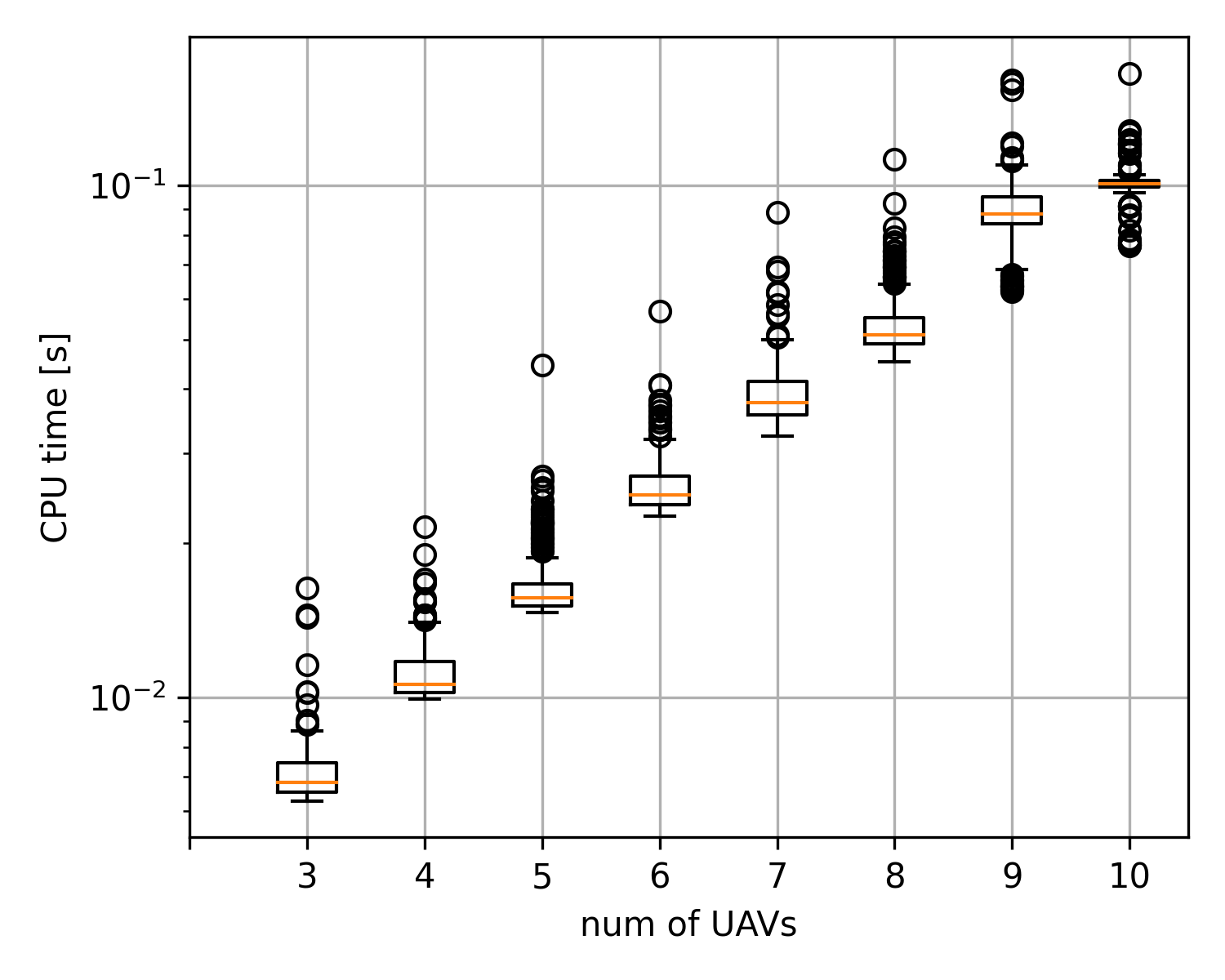}
    \caption{Number of drones versus CPU time for generating a predicted trajectory. The tested CPU is Intel Core i7-10750H, 2.6Hz. Note that the vertical axis is in log scale.} 
    \label{fig:CPU_time}
\end{figure}

With the same initial condition and reference pose, we test the case with various numbers of UAVs. Fig.~\ref{fig:CPU_time} shows the box plot of total CPU elapsed time to generate a predicted trajectory. Since the proposed NMPC is a centralized approach, the number of optimization parameters and constraints increases with the number of units. This leads to an exponential growth of CPU time as the number of UAVs increases. Be that as it may, the medium CPU time is still less than 100\,ms even 10~UAVs are involved, which clearly indicates the viability of the proposed NMPC in real-time and operations in the field.
\section{Discussions}
\label{eq:discussions}
The proposed NMPC regards cable tensions as inputs instead of state variables.
Thus no sensors are needed to measure cable tensions.
However, this controller does need full states of the load and cable directions, especially the angular acceleration of the cable direction, which can be challenging to obtain.
A potentially practical way to estimate load and cable states in real-world operations is to equip a sensor on each UAV to measure the cable information, including cable direction, angular rate and acceleration with respect to the UAVs.
Then the load states can be indirectly obtained.
In this work, we neglect all uncertainties on models and state estimations, which need to be thoroughly investigated in future work.

In the simulation validation, we demonstrate that with the proposed NMPC it is possible to address multiple constraints to guarantee the operation safety.
Even though the system dynamic and several inequality constraints are nonlinear, the ACADOS solver can still run NMPC at average computation time of of 100\,ms on a laptop, in the worst case of large system composed by 10~UAVs.
However, despite the consideration of the collective thrust limits of each UAV, the proposed NMPC does not consider the limits of the individual actuator of each UAV.
Theoretically, it is possible to formulate the NMPC to address these limits using inequality path constraints.
But these constraints will be too complex to be dealt with in a real-time manner using existing solvers.
A practical solution to address these constraints, as is proposed in this paper, is to use an inner-loop controller.
As is pointed out in~\cite{sreenath2013dynamics}, another option is to plan a reference trajectory that minimizes the 6\,th derivative of the load pose that leads to a minimum snap trajectory for the UAVs.
\section{Conclusion}
\label{sec:conclusion}
This paper introduces a centralized nonlinear model predictive control method for full-pose manipulation of a cable-suspended load using multiple UAVs. 
We have validated the control algorithm through numerical simulations to quickly track a pose reference while complying with multiple safety-related constraints. 
Future works will include analysis of parameter uncertainties and real-world experimental validations.

\section*{Acknowledgement}
We thank Dr. Chiara Gabellieri for helping us testing the software.

\printbibliography

@inproceedings{gassner2017dynamic,
  title={Dynamic collaboration without communication: Vision-based cable-suspended load transport with two quadrotors},
  author={Gassner, Michael and Cieslewski, Titus and Scaramuzza, Davide},
  booktitle={2017 IEEE International Conference on Robotics and Automation (ICRA)},
  pages={5196--5202},
  year={2017},
  organization={IEEE}
}

@inproceedings{tagliabue2017collaborative,
  title={Collaborative transportation using mavs via passive force control},
  author={Tagliabue, Andrea and Kamel, Mina and Verling, Sebastian and Siegwart, Roland and Nieto, Juan},
  booktitle={2017 IEEE International Conference on Robotics and Automation (ICRA)},
  pages={5766--5773},
  year={2017},
  organization={IEEE}
}

@article{tognon2018aerial,
  title={Aerial co-manipulation with cables: The role of internal force for equilibria, stability, and passivity},
  author={Tognon, Marco and Gabellieri, Chiara and Pallottino, Lucia and Franchi, Antonio},
  journal={IEEE Robotics and Automation Letters},
  volume={3},
  number={3},
  pages={2577--2583},
  year={2018},
  publisher={IEEE}
}

@inproceedings{sundin2022decentralized,
  title={Decentralized Model Predictive Control for Equilibrium-based Collaborative UAV Bar Transportation},
  author={Sundin, Roberto C and Roque, Pedro and Dimarogonas, Dimos V},
  booktitle={2022 International Conference on Robotics and Automation (ICRA)},
  pages={4915--4921},
  year={2022},
  organization={IEEE}
}

@article{foehn2017fast,
  title={Fast trajectory optimization for agile quadrotor maneuvers with a cable-suspended payload},
  author={Foehn, Philipp and Falanga, Davide and Kuppuswamy, Naveen and Tedrake, Russ and Scaramuzza, Davide},
  year={2017},
  publisher={Robotics: Science and Systems Foundation}
}

@inproceedings{guerrero2015passivity,
  title={Passivity based control for a quadrotor UAV transporting a cable-suspended payload with minimum swing},
  author={Guerrero, Maria Eusebia and Mercado, DA and Lozano, Rogelio and Garc{\'\i}a, CD},
  booktitle={2015 54th IEEE Conference on Decision and Control (CDC)},
  pages={6718--6723},
  year={2015},
  organization={IEEE}
}

@inproceedings{son2018model,
  title={Model predictive control of a multi-rotor with a suspended load for avoiding obstacles},
  author={Son, Clark Youngdong and Seo, Hoseong and Kim, Taewan and Kim, H Jin},
  booktitle={2018 IEEE International Conference on Robotics and Automation (ICRA)},
  pages={5233--5238},
  year={2018},
  organization={IEEE}
}

@article{goodarzi2015geometric,
  title={Geometric control of a quadrotor UAV transporting a payload connected via flexible cable},
  author={Goodarzi, Farhad A and Lee, Daewon and Lee, Taeyoung},
  journal={International Journal of Control, Automation and Systems},
  volume={13},
  pages={1486--1498},
  year={2015},
  publisher={Springer}
}

@article{tartaglione2017model,
  title={Model predictive control for a multi-body slung-load system},
  author={Tartaglione, Gaetano and D’Amato, Egidio and Ariola, Marco and Rossi, Pierluigi Salvo and Johansen, Tor Arne},
  journal={Robotics and Autonomous Systems},
  volume={92},
  pages={1--11},
  year={2017},
  publisher={Elsevier}
}

@inproceedings{zurn2016mpc,
  title={MPC controlled multirotor with suspended slung load: System architecture and visual load detection},
  author={Z{\""u}rn, Markus and Morton, Kye and Heckmann, Alexander and McFadyen, Aaron and Notter, Stefan and Gonzalez, Felipe},
  booktitle={2016 IEEE Aerospace conference},
  pages={1--11},
  year={2016},
  organization={IEEE}
}

@mastersthesis{moreno2018planning,
  title={Planning and control of a multiple-quadcopter system cooperatively carrying a slung payload in dynamical environments. A centralized model predictive control solution},
  author={Moreno Caireta, I{\~n}igo},
  type={{B.S.} thesis},
  year={2018},
  school={Universitat Polit{\`e}cnica de Catalunya}
}

@inproceedings{lee2013geometric,
  title={Geometric control of cooperating multiple quadrotor UAVs with a suspended payload},
  author={Lee, Taeyoung and Sreenath, Koushil and Kumar, Vijay},
  booktitle={52nd IEEE conference on decision and control},
  pages={5510--5515},
  year={2013},
  organization={IEEE}
}

@inproceedings{de2019flexible,
  title={Flexible collaborative transportation by a team of rotorcraft},
  author={De Marina, Hector Garcia and Smeur, Ewoud},
  booktitle={2019 International Conference on Robotics and Automation (ICRA)},
  pages={1074--1080},
  year={2019},
  organization={IEEE}
}

@article{bernard2011autonomous,
  title={Autonomous transportation and deployment with aerial robots for search and rescue missions},
  author={Bernard, Markus and Kondak, Konstantin and Maza, Ivan and Ollero, Anibal},
  journal={Journal of Field Robotics},
  volume={28},
  number={6},
  pages={914--931},
  year={2011},
  publisher={Wiley Online Library}
}

@article{lee2017geometric,
  title={Geometric control of quadrotor UAVs transporting a cable-suspended rigid body},
  author={Lee, Taeyoung},
  journal={IEEE Transactions on Control Systems Technology},
  volume={26},
  number={1},
  pages={255--264},
  year={2017},
  publisher={IEEE}
}

@inproceedings{wehbeh2020distributed,
  title={Distributed model predictive control for UAVs collaborative payload transport},
  author={Wehbeh, Jad and Rahman, Shatil and Sharf, Inna},
  booktitle={2020 IEEE/RSJ International Conference on Intelligent Robots and Systems (IROS)},
  pages={11666--11672},
  year={2020},
  organization={IEEE}
}

@article{sanalitro2020full,
  title={Full-pose manipulation control of a cable-suspended load with multiple UAVs under uncertainties},
  author={Sanalitro, Dario and Savino, Heitor J and Tognon, Marco and Cort{\'e}s, Juan and Franchi, Antonio},
  journal={IEEE Robotics and Automation Letters},
  volume={5},
  number={2},
  pages={2185--2191},
  year={2020},
  publisher={IEEE}
}

@inproceedings{manubens2013motion,
  title={Motion planning for 6-D manipulation with aerial towed-cable systems},
  author={Manubens, Montserrat and Devaurs, Didier and Ros, Lluis and Cort{\'e}s, Juan},
  booktitle={Robotics: science and systems (RSS)},
  pages={8p},
  year={2013}
}

@article{li2021cooperative,
  title={Cooperative transportation of cable suspended payloads with MAVs using monocular vision and inertial sensing},
  author={Li, Guanrui and Ge, Rundong and Loianno, Giuseppe},
  journal={IEEE Robotics and Automation Letters},
  volume={6},
  number={3},
  pages={5316--5323},
  year={2021},
  publisher={IEEE}
}

@inproceedings{erunsal2022linear,
  title={Linear and Nonlinear Model Predictive Control Strategies for Trajectory Tracking Micro Aerial Vehicles: A Comparative Study},
  author={Erunsal, IK and Zheng, Jianhao and Ventura, Rodrigo and Martinoli, Alcherio},
  booktitle={2022 IEEE/RSJ International Conference on Intelligent Robots and Systems (IROS)},
  pages={12106--12113},
  year={2022},
  organization={IEEE}
}

@inproceedings{fresk2013full,
  title={Full quaternion based attitude control for a quadrotor},
  author={Fresk, Emil and Nikolakopoulos, George},
  booktitle={2013 European control conference (ECC)},
  pages={3864--3869},
  year={2013},
  organization={IEEE}
}

@inproceedings{sreenath2013dynamics,
  title={Dynamics, control and planning for cooperative manipulation of payloads suspended by cables from multiple quadrotor robots},
  author={Sreenath, Koushil and Kumar, Vijay},
  booktitle={Robotics: science and systems (RSS)},
  pages={8p},
  year={2013}
}

@inproceedings{falanga2018pampc,
  title={PAMPC: Perception-aware model predictive control for quadrotors},
  author={Falanga, Davide and Foehn, Philipp and Lu, Peng and Scaramuzza, Davide},
  booktitle={2018 IEEE/RSJ International Conference on Intelligent Robots and Systems (IROS)},
  pages={1--8},
  year={2018},
  organization={IEEE}
}

@article{tal2020accurate,
  title={Accurate tracking of aggressive quadrotor trajectories using incremental nonlinear dynamic inversion and differential flatness},
  author={Tal, Ezra and Karaman, Sertac},
  journal={IEEE Transactions on Control Systems Technology},
  volume={29},
  number={3},
  pages={1203--1218},
  year={2020},
  publisher={IEEE}
}

@article{sun2022comparative,
  title={A comparative study of nonlinear mpc and differential-flatness-based control for quadrotor agile flight},
  author={Sun, Sihao and Romero, Angel and Foehn, Philipp and Kaufmann, Elia and Scaramuzza, Davide},
  journal={IEEE Transactions on Robotics},
  volume={38},
  number={6},
  pages={3357--3373},
  year={2022},
  publisher={IEEE}
}

@inproceedings{lee2010geometric,
  title={Geometric tracking control of a quadrotor UAV on SE (3)},
  author={Lee, Taeyoung and Leok, Melvin and McClamroch, N Harris},
  booktitle={49th IEEE conference on decision and control (CDC)},
  pages={5420--5425},
  year={2010},
  organization={IEEE}
}

@article{verschueren2022acados,
  title={acados—a modular open-source framework for fast embedded optimal control},
  author={Verschueren, Robin and Frison, Gianluca and Kouzoupis, Dimitris and Frey, Jonathan and Duijkeren, Niels van and Zanelli, Andrea and Novoselnik, Branimir and Albin, Thivaharan and Quirynen, Rien and Diehl, Moritz},
  journal={Mathematical Programming Computation},
  volume={14},
  number={1},
  pages={147--183},
  year={2022},
  publisher={Springer}
}

@article{frison2020hpipm,
  title={HPIPM: a high-performance quadratic programming framework for model predictive control},
  author={Frison, Gianluca and Diehl, Moritz},
  journal={IFAC-PapersOnLine},
  volume={53},
  number={2},
  pages={6563--6569},
  year={2020},
  publisher={Elsevier}
}

@article{diehl2002real,
  title={Real-time optimization and nonlinear model predictive control of processes governed by differential-algebraic equations},
  author={Diehl, Moritz and Bock, H Georg and Schl{\"o}der, Johannes P and Findeisen, Rolf and Nagy, Zoltan and Allg{\"o}wer, Frank},
  journal={Journal of Process Control},
  volume={12},
  number={4},
  pages={577--585},
  year={2002},
  publisher={Elsevier}
}

@article{bock1984multiple,
  title={A multiple shooting algorithm for direct solution of optimal control problems},
  author={Bock, Hans Georg and Plitt, Karl-Josef},
  journal={IFAC Proceedings Volumes},
  volume={17},
  number={2},
  pages={1603--1608},
  year={1984},
  publisher={Elsevier}
}

@article{sun2019aerodynamic,
  title={Aerodynamic model identification of a quadrotor subjected to rotor failures in the high-speed flight regime},
  author={Sun, Sihao and de Visser, Coen},
  journal={IEEE Robotics and Automation Letters},
  volume={4},
  number={4},
  pages={3868--3875},
  year={2019},
  publisher={IEEE}
}

\end{document}